\def\year{2019}\relax
\documentclass[letterpaper]{article} 
\usepackage{aaai19}  
\usepackage{times}  
\usepackage{helvet}  
\usepackage{courier}  
\usepackage{url}  
\usepackage{graphicx}  
\usepackage{stfloats}
\usepackage{latexsym}
\usepackage{xcolor}
    \definecolor{red}{HTML}{E51400} 
    \definecolor{blue}{HTML}{0050EF} 
    \definecolor{green}{HTML}{008A00} 
    \definecolor{purple}{HTML}{AA00FF} 
    \definecolor{orange}{HTML}{FF7F00}
    \definecolor{gray}{HTML}{848482} 
\usepackage{soul}
\usepackage[utf8]{inputenc}
\usepackage[small]{caption}
\usepackage{amsmath}
\usepackage{amssymb}
\usepackage{hyperref}
\hypersetup{colorlinks=true, linkcolor=red, citecolor=blue, anchorcolor=blue, urlcolor=blue}  
\usepackage{bm}
\begin{document}
%
\title{Modeling Local Dependence in Natural Language with \\
	Multi-Channel Recurrent Neural Networks}

\author{Chang Xu$^1$, Weiran Huang$^2$, Hongwei Wang$^3$, Gang Wang$^4$ and Tie-Yan Liu$^5$\\
$^{1,4}$College of Computer Science, Nankai University, \{changxu, wgzwp\}@nbjl.nankai.edu.cn\\
$^2$Tsinghua University, huang.inbox@outlook.com\\
$^3$Shanghai Jiao Tong University, wanghongwei55@gmail.com\\ $^5$Microsoft Research, tie-yan.liu@microsoft.com
}
\maketitle
\begin{abstract}
Recurrent Neural Networks (RNNs) have been widely used in processing natural language tasks and achieve huge success.
Traditional RNNs usually treat each token in a sentence uniformly and equally. However, this may miss the rich semantic structure information of a sentence,
which is useful for understanding natural languages.
Since semantic structures such as word dependence patterns are not parameterized, it is a challenge to capture and leverage structure information.
In this paper, we propose an improved variant of RNN, Multi-Channel RNN (MC-RNN), to dynamically capture and leverage local semantic structure information.
Concretely, MC-RNN contains multiple channels, each of which represents  a local dependence pattern at a time.
An attention mechanism is introduced to combine these patterns at each step, according to the semantic information.
Then we parameterize structure information by adaptively selecting the most appropriate connection structures among channels.
In this way, diverse local structures and dependence patterns in sentences can be well captured by MC-RNN.
To verify the effectiveness of MC-RNN, we conduct extensive experiments on typical natural language processing tasks, including neural machine translation, abstractive summarization, and language modeling.
Experimental results on these tasks all show significant
improvements of MC-RNN over current top systems.
\end{abstract}

\section{Introduction}
Recurrent neural networks (RNNs), designed with recurrent units and parameter sharing, have demonstrated outstanding ability in modeling sequential data and achieved success in various Nature Language Processing (NLP) tasks, such as language modeling \cite{merity2017regularizing}, machine translation
\cite{bahdanau2016actor,huang2017neural}, abstractive summarization \cite{suzuki2017cutting}, and dialog systems \cite{asri2016sequence}.
Traditional RNNs
produce hidden state vectors one by one through recurrent computations, treating all tokens in the sequence uniformly and equally. This would limit the applicability of the model.

Notice that the meaning of a sentence is determined by two main factors: the meaning of each word and the rule of combining
them.
Thus, the semantic structure information is essential for understanding texts.
In fact, natural languages exhibit strong local structures in terms of semantics. For example, in sentence ``\textit{We must find the missing document at all costs.}'', word ``\textit{at}'',``\textit{all}'' and ``\textit{cost}'' form a phrase and are strongly related to each other, so do word ``\textit{missing}'' and ``\textit{document}''.
In contrast,  ``\textit{document}'' and ``\textit{at}" have no such strong semantic correlation, although they are also neighbors.
This indicates that phrase structures are very important and essential for capturing a wealth of semantic information and understanding the meaning of sentences.
However, they are hard to be modeled by traditional RNNs in the uniformly sequential way.

To capture and leverage the semantic structure information, there are two main challenges. First, since there are diverse word dependence patterns, flexible and learnable structure modeling method is preferred than predefined connections or fixed topology. Second, the local structures and word dependence patterns in sentences are discrete symbols rather than regular learnable model parameters, so it is non-trivial to capture and parameterize them.

In this paper, we propose Multi-Channel RNN (MC-RNN), a novel RNN structure that makes full use of local structure information and dependency patterns of natural language, without requiring additional grammar knowledge.
To tackle the first challenge, MC-RNN enumerates possible local structure patterns in a sentence. Different from the simple dependence among hidden states in conventional RNNs (i.e., a hidden state connects directly only to its immediate predecessor), MC-RNN is designed with multiple channels, each channel is partitioned into multiple blocks. With such a multi-block design, a hidden state takes the outputs of several direct/indirect predecessors as inputs. Different partitions of the nodes in different channels focus on different local structures at each step. Together, many possible local dependency patterns can be enumerated. Compared with the methods of predefined connections or fixed topology, our proposed MC-RNN with multiple channels and multiple blocks is more representative in modeling diverse patterns, and flexible to RNN variants.

For the second challenge, we propose an
attention mechanism among channels to select proper local structure or dependency patterns at each step. In this way, the structures and connections in sentences can be parameterized, and the discrete local structures and dependency patterns can be mined and transformed into a regular learnable problem. Concretely, an attention mechanism among channels is proposed to aggregate the outputs of all channels with different weights according to the sentence semantic information. The channels which more accurately capture local structures/dependence at the current step are expected to be assigned larger weight. As a result, we can dynamically and adaptively select the most appropriate connection structure among different channels.

Our main contributions are summarized as follows.
\begin{itemize}
    \item
    We propose MC-RNN, a novel varient of RNN with the multi-channel multi-block design, to learn and capture local patterns/dependence in sequential text data. By enumerating different structures in multi-channels, MC-RNN parameterizes the structure learning problem.
    \item
    We introduce an attention mechanism among channels which can aggregate the outputs of all channels with different weights according to sentence semantic information. The most appropriate connection structure among different channels can be dynamically and adaptively selected at the different steps.
    \item
    We apply MC-RNN to three NLP tasks: machine translation, language modeling, and abstractive summarization.
    The experimental results show that MC-RNN significantly outperforms previous works.

\end{itemize}

\section{Related Work}
Our work is related to previous studies about learning sentence representations considering structures information to improve the performance of NLP tasks~\cite{kim2017structured,daniluk2017frustratingly,liu2017learning,chung2016hierarchical,Koutn2014A,soltani2016higher,Wang2016Recurrent}. The related work can mainly be classified into three categories.

The methods in the first category are based on external knowledge such as grammatical knowledge \cite{su2017lattice}, sentence structure information \cite{zhu2015long,DBLP:conf/acl/TaiSM15,maillard2017jointly} and surrounding context information~\cite{liu2017learning}.
However, external knowledge usually can only be used in the encoder of sequence-to-sequence models rather than the decoder.
A complete sentence should be provided as input to use grammatical knowledge, however, this does not hold for the target sequence, since words will be generated by the decoder.
In addition, prior knowledge may be inaccurate and noisy in practice.

The second category of methods intends to design new topological structures of RNN \cite{soltani2016higher,Wang2016Recurrent}. For example, Higher Order RNN \cite{soltani2016higher} connects more preceding states to the current state to better explore the local dependence. However, the topological connections are predefined and keep fixed in the model, which limits its ability in discovering a wide range of diverse local structures.

The third category of methods attempts to design new recurrent computation functions in RNN \cite{chung2016hierarchical,Koutn2014A}. For example, HM-RNN \cite{chung2016hierarchical} adaptively learns hard boundaries among input tokens, which enhances information flow by reducing the update frequencies of high abstractive levels. It achieves high flexibility in determining whether to keep/update the hidden state of previous time step unchanged. However, the information flow may be broken between two boundaries when the hidden states are not updated within a semantic unit. Furthermore, HM-RNN is difficult to be adapted to other types of RNNs since it changes the recurrent computation of RNN units.

There are some other studies, which are not based on recurrent structures, also model word relations to improve model performance, including convolutional neural network (CNN) ~\cite{gehring2017convolutional} and transformer~\cite{vaswani2017attention}. However, since they are not recurrent models, their ability to model ordering information is not as strong as RNN. Specifically, recent studies \cite{yang2018unsupervised} show that transformer model suffers from the lost of temporal order information, especially within an attention, which is very important for modeling structures in the sentence. CNN suffers from similar problem of losing temporal order information, and the fixed kernel size also limits it's flexibility.

We remark that our proposed MC-RNN is different from existing methods and combines their advantages:
(1) MC-RNN requires no external grammar knowledge as supervisory information.
(2) By forming locally connections among adjacent input units, MC-RNN can model diverse dependence patterns in sequences and better leverage rich information of adjacent semantic units.
(3) It is highly flexible and can be easily adapted to any RNN variants.

\section{Model Description}

In this section, we introduce the MC-RNN model. Figure 1 shows an example of  the MC-RNN layer with 3 channels. MC-RNN consists of multiple channels, and each channel consists of multiple blocks. All blocks in a channel are joined, head-to-end, in a single line, thus the sequential property and temporal order information can be retained. Each node is connected with all previous nodes in the same block, which strengthens local dependence for the nodes within the block. 
We set the blocks of neighboring channels has one step staggered with each other in a progressive way, so that all possible local patterns/dependence whose length is no more than the block size can be covered. 
At each step, all channels in the layer are aggregated by an attention module, which makes it possible to dynamically and adaptively select the most appropriate local structure among different channels at the different step.

We will introduce the details of multiple-channel design in Section~\ref{multichannel}  and the attention module in Sectionn~\ref{attention}. Moreover, we leave the discussion of how to determine the block size in Section~\ref{blocksize} and the time cost of MC-RNN in Section~\ref{timecost}.

\begin{figure*}[ht!]
    \centering
    \includegraphics[page=1, width=0.9\linewidth]{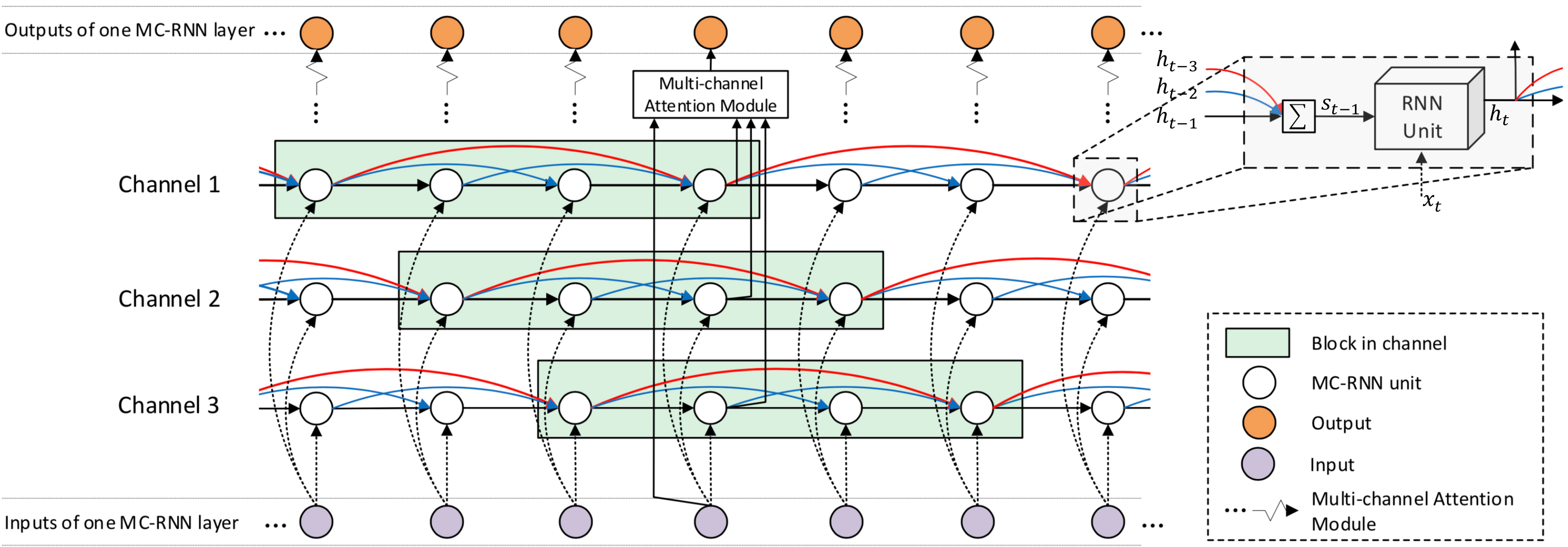}
    \vskip -0.5em
    \caption{Illustration of the structure of one-layer MC-RNN with 3 channels. Each channel in the MC-RNN layer contains several blocks.  Local connections are built in each block. 
Solid lines with the same color (red/blue/black) share the same parameter matrices.
Parameters in different channels are all shared with each other. The outputs of hidden states from of all channels are fed into the multichannel attention module to generate an aggregated hidden state to next layer at each step. Channels can be computed in parallel (see Section \ref{timecost}). Best viewed in color.}
    \label{general}
    \vskip -1em
\end{figure*}

\subsection{Capturing Rich Patterns with Multiple Channels}\label{multichannel}

To ensure the ability to represent most of local structure patterns for MC-RNN, we design different connection mechanism for different channels. As shown in Figure 1, channels are different from each other in terms of the partition of blocks. The nodes in the same block are fully connected. For example, if phrase ``\textit{at all costs}'' is in the block, word ``\textit{costs}'' will be both connected to word ``\textit{at}'' and ``\textit{all}''. Such connection mechanism makes MC-RNN have inherent advantage to model local structure information than traditional RNN. 
Since the begin-to-end block composition in one channel can only handle part of the possible local structure patterns, we set the blocks of neighboring channels has one step staggered with each other in a progressive way. In this way, all possible local structures or dependency patterns whose length is no more than the block size can be enumerated.

\begin{figure*}[t]
    \centering
    \begin{minipage}[t]{1\linewidth}
        \centering
        \includegraphics[page=1, width=0.9\linewidth]{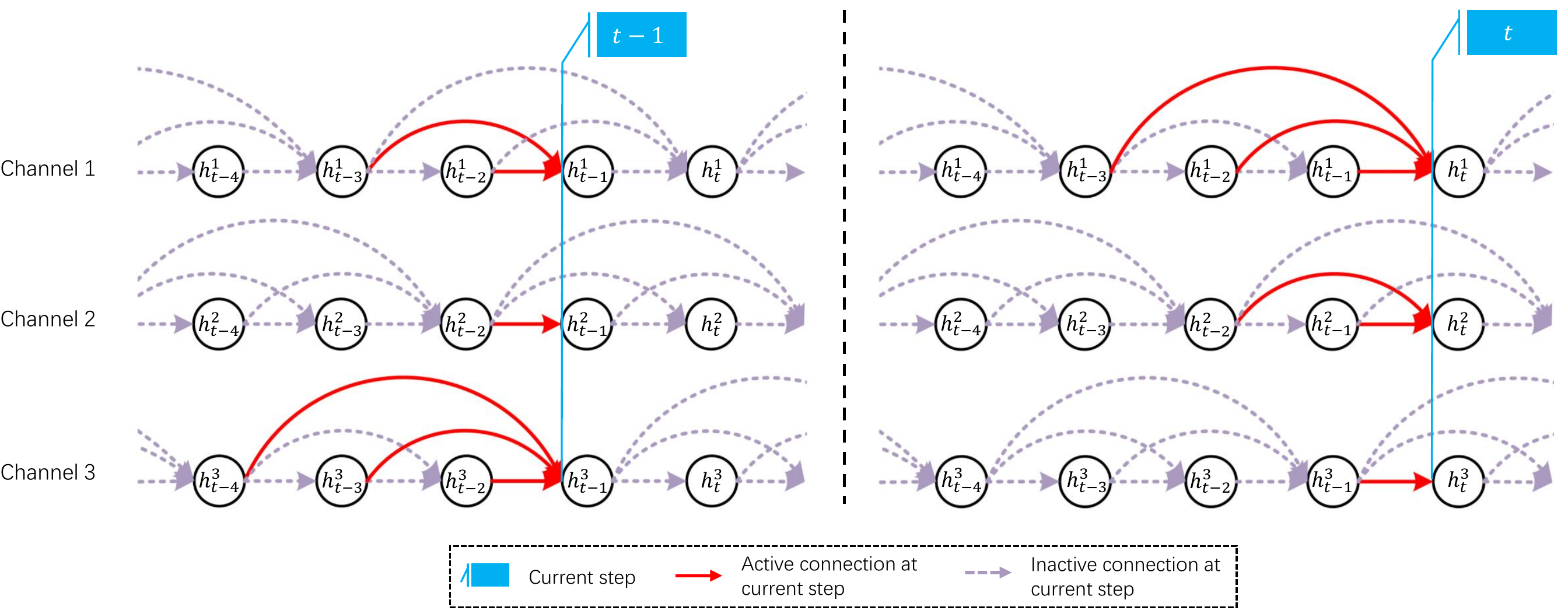}
        \caption{
Dependence patterns in different channels at time step $t-1$ and $t$. The red lines indicates active connections at current time step. At time step $t-1$, the red lines in channel $1, 2, 3$ represent 3-word/ 2-word/ 4-word dependence patterns respectively, while at time step $t$, the red lines in channel $1, 2, 3$ represent 4-word/ 3-word/ 2-word dependence patterns respectively. For example, channel 2 at time step $t$ represents the dependence pattern that the current word is strongly dependent on previous two words.}
        \label{fig:dependency}
    \end{minipage}
\end{figure*}

Figure \ref{fig:dependency} illustrates the detail of channels inside an MC-RNN. To be clear, we list all definitions of used symbols in Table~\ref{symbols}. Suppose there are n channels in the MC-RNN. In each channel $k\in[n]$, the first block consists units $h_{2-k}^k, h_{3-k}^k, \cdots, h_{n+1-k}^k$. 
We define $h_{t}^k$ to be zero vector for $t\le 0$, namely, we pad zero vectors for the first block for all the channels except the first one. 
With full connections in each block, the inputs of each recurrent unit include not only its immediate predecessor but also from the historical units within a certain distance. Thus, MC-RNN can capture a strong dependence between words in a phrase, and make compact representations for the phrase.

\begin{table}[t]
	\centering
	\caption{The definitions of important symbols used in our paper.}
	\begin{tabular}{l|l}
		\hline
		Symbol   & Definition   \\ \hline\hline
		$n$  & block size, i.e. number of nodes in a block \\
		$n-1$  & number of channels in each MC-RNN layer  \\ 
		$k$ & index of channel in one MC-RNN layer     \\ 
		$t$ & index of time step \\
		$m_t^{k}$ & in-degree of a node, i.e. how many precessing\\  
		 & nodes are connected to current node at step\\
		 & $t$ in channel ${k}$ \\ 
		$x_t$ & the input from previous layer at step $t$-th \\
		$h_t^k$ & output of the node at step $t$ in channel $k$\\
		$s_{t-1}^k$ & temporal input at step $t$ in channel $k$ \\
        \hline
	\end{tabular}
	\label{symbols}
\end{table}
We use pair notation $(t,k)$ to denote the node at step $t$ in channel $k$ and $m_t^k$ to denote the in-degree of node $(t,k)$, i.e.,
the number of predecessors connected to node $(t,k)$. Thus, we have
\begin{equation}\label{eq1}
    m^k_t = (t-k-1) \ {\rm{mod}} \ (n-1) + 1.
\end{equation}

We use $f\colon \mathbb{R}^{N_h}\times\mathbb{R}^{N_x} \rightarrow \mathbb{R}^{N_h}$ to represent the recurrent computation in a traditional RNN,
where $N_h$ is the dimension of hidden state and $N_x$ is the dimension of RNN input.
Thus, the output of the node at any step $t$, denoted as $\bm{h}_t$, can be computed by $\bm{h}_t = f(\bm{s}_{t-1}, \bm{x}_t)$, where $\bm{s}_{t-1}$ is the temporal input at step $t$ and $\bm{x}_t$ is the input from previous layer at step $t$.
Since MC-RNN  does not require any modifications to the traditional RNN formulation, $f$ could be any recurrent function used in vanilla RNN, GRU, LSTM, etc. In traditional RNNs, $\bm{s}_{t-1}$ is the same as $\bm{h}_{t-1}$, while in the MC-RNN, we define the temporal input at step $t$ in channel $k$ as
\begin{equation}
    \bm{s}_{t-1}^k = \dfrac{1}{m^k_t}  \sum^{m^k_t}_{j=1}  W_j \bm{h}^k_{t-j},
    \label{stk}
\end{equation}
where $m_t^k$ is computed by Eq.~\eqref{eq1} and $W_j$ is the weight matrix between node $(t,k)$ and node $(t-j,k)$.
Note that the weight matrix between two nodes only depends on their distance, i.e., weight matrices of solid lines with the same color in Figure~\ref{general} are the same. 
In other words, we compute the  temporal input of each recurrent computation by taking a weighted average of previous outputs within the block.
Then we apply the recurrent computation $f$ to get the output:
\begin{equation}
    \bm{h}_t^k = f(\bm{s}^k_{t-1}, \bm{x}_t).
\end{equation}
Note that learnable parameters including RNN internal parameters and weights in blocks are shared among different channels.

\subsection{Aggregating Patterns by an Attention Module}\label{attention}
Since each channel of MC-RNN is designed to have different topological connections representing different dependence patterns, we propose an attention mechanism to combine them by dynamically adjusting the weight of each channel.
For example, in the case of a three-word phrase, when we process the third word, we want to make use of the information of the first two terms explicitly because they belong to the same phrase and have local semantic coherence.
Therefore, we want the channel that directly connects the word with its two predecessors to have the largest weight.

Specifically, we use the attention mechanism to obtain the weighted average of each channel's hidden state at time $t$ as the input to next layer, which is denoted as

\begin{equation}
 \bm{h}^{att}_t = \sum^n_{k=1} \alpha^k_t \bm{h}^k_t,
\end{equation}
where $\alpha^k_t$ is the attention weight for the $k$-th channel at time $t$.
The attention weight $\alpha^k_t$ is calculated by
\begin{equation}
 \alpha^k_t = \frac{ {\exp(e^k_t)}}{ \sum^n_{i=1} {\exp(e^i_t) }},
\end{equation}
and $e^k_t$ is defined as
\begin{equation}
 e^k_t = \bm{r}^T   \  {\rm{tanh}} \left( {V\cdot \begin{bmatrix}\bm{h}_t^k \\ \bm{x}_t\end{bmatrix}} \right).
\end{equation}
where $\bm{r} \in \mathbb{R} ^{N_h}$ and $V \in \mathbb{R} ^{N_h \times (N_h+N_x)}$ are weight matrices of the attention module. The combination weights are determined by the current states of the channels as well as the input from previous layer. In this way, local semantics are taken into consideration when integrating the output of different topical connections and diverse dependency patterns are modeled.

\section{Experiments}

In this section, we evaluate the performance of MC-RNN on three different tasks, including neural machine translation, abstractive summarization and language modeling. Both neural machine translation and abstractive summarization are sequence-to-sequence text generation tasks. 
They first try to understand the source sentence using an encoder and then generate a target sentence. The process of sentences understanding and generating would both benefit from capturing the local dependence patterns and structure information in sentences. Besides, language modeling task aims at predicting next words conditioned on previous words, which would intuitively rely more on the local information. 
We also study the learned dependence patterns and analyze the performance of our model with different sentence lengths in this section.



\subsection{Machine Translation}
\subsubsection{Experimental Setups}
The data we use is the German-English (De-En for short) machine translation track of the IWSLT 2014 evaluation campaign \cite{cettolo2014report}, which is popular in machine translation community \cite{bahdanau2016actor,ranzato2015sequence}. We follow the same pre-processing as described in above works. The training/dev/test dataset respectively contains about 153$k$/7$k$/7$k$ De-En sentences pairs. We pre-processed the corpus with byte pair encoding (BPE) \cite{sennrich2015neural}, since BPE has been shown to be an effective approach to handle the large vocabulary issue in NMT and thus has better performance than word-based vocabulary. Following the settings of previous works, we extract about 25k sub-word tokens as vocabulary. BLEU \cite{papineni2002bleu} is used as the evaluation metric. 

We first implement a basic baseline model following the most widely used sequence to sequence framework RNNSearch \cite{bahdanau2014neural}, denoted as  Baseline-RNN. And then we build MC-RNN following the settings of Baseline-RNN but adding block connections and multi-channel attention. The performance of MC-RNN is compared to that of Baseline-RNN so that we can make a direct comparison and see the improvement brought by our method. The encoders and decoders of our model and Baseline-RNN are all equipped with 2-layer LSTM with word embedding size 256 and hidden state size 256. BPE is used as pre-processing for MC-RNN and baselines. For Baseline-RNN, all hyperparameters such as dropout ratio and gradient clipping threshold are chosen via cross-validation on the dev set. For our MC-RNN, we followed the settings of Baseline-RNN, and keep all hyperparameters the same with Baseline-RNN. During training, we automatically halve the learning rate according to validation performance on dev set and stop when the performance is not improved any more. For decoding, we use beam search~\cite{sutskever2014sequence} with width 5.

We also compared our method with two recent models, which are representative of two classes of studies to capture the local structure of sentences without external knowledge. One is HO-RNN \cite{soltani2016higher} which changes the topological structure of RNN. The other one is HM-RNN \cite{chung2016hierarchical} which modifies the recurrent computations.
In addition, we compare our model with the following RNN based methods on this task.
(1) Actor-critic \cite{bahdanau2016actor}, an approach to training neural networks to generate sequences using reinforcement learning.
(2) NPMT-LM \cite{huang2017neural}, a neural phrase-based machine translation system that models phrase structures
in the target language.

\subsubsection{Experimental Results}
\label{MT}\label{blocksize}

We first study the effect of the number of channels $k$ and block size $n$ with $n=k+1$. In Table \ref{MNT}, we examine the performance of different model structures. MC-RNN-$k$ stands for MC-RNN with $k$ channels, which varies from 2 to 4. 
The results show that MC-RNN-3 achieves significantly better performance than MC-RNN-2, which matches our intuition that a model with larger block size has stronger ability to learn complex dependence patterns. However, when we further increase the number of channels from 3 to 4, MC-RNN-4 performs worse than MC-RNN-3 due to the larger model size of MC-RNN-4. Since there are too many parameters to learn, the optimization process would be more complex and prone to overfitting. 

The experiment results from Table \ref{MNT} show that MC-RNN achieves 1.20 BLEU gains over Baseline-RNN model. Furthermore, MC-RNN (32.23) even outperforms the best result among all previous RNN based methods,  31.29 of HO-RNN, by 0.94.

\begin{table}[t]
	\centering
	\caption{BLEU scores on IWLST 2014 De-En dataset.}
	\begin{tabular}{l|r|rr}
		\hline
		Methods   &Params & BLEU   \\ \hline\hline
		Actor-critic  &-& 28.53 \\
		NPMT-LM    &-& 29.16  \\ 
		HM-RNN &25$M$& 30.60     \\ 
		HO-RNN &30$M$& 31.29 \\
		Baseline-RNN &25$M$& 31.03 \\ \hline
		MC-RNN-2 &28$M$&31.98 \\
		MC-RNN-3 &29$M$& $\star$32.23 \\
		MC-RNN-4 &31$M$& 32.09 \\  \hline
	\end{tabular}
	\label{MNT}
\end{table}

\subsection{Abstractive Summarization}
\subsubsection{Experimental Setups}
The dataset we use is Gigaword corpus \cite{GraffEnglish},
which consists of headline-article pairs.
The task is to generate the headline of the given article.
We pre-process the dataset similar to \cite{rush2015neural}, resulting in 3.8\textit{M} training article-headline pairs, 190\textit{k} for validation and 2000 for test. 
Following the settings in \cite{shen2016neural}, we set the vocabulary size to 30\textit{k}.
ROUGE F1 score \cite{Flick2004ROUGE} is used as evaluation criterion for summarization task.

On this task, MC-RNN also follows the settings of Baseline-RNN with LSTM as the recurrent unit. Both the encoder and the decoder have 4 layers. The embedding size of our model is 256, and the LSTM hidden state size is 256. The mini-batch size is 64 and the learning rate is halved when the dev performance stops increasing. Similar to the machine translation task, we also use HM-RNN and HO-RNN with the same configuration of Baseline-RNN as baselines. For decoding, we use the beam search with width 10 which is the same with previous works.

\subsubsection{Experimental Results}

From Table \ref{ABS:RNNsearch}, we have similar observations to the above machine translation experiments that MC-RNN with 3 channels achieves the best performance.
All of the three MC-RNN models perform better than Baseline-RNN by non-trivial margins, demonstrating the effectiveness of our method.
Our best model, MC-RNN-3 achieves 1.90, 1.45, 1.48 points improvement compared to the Baseline-RNN model on unigram based ROUGE-1, bigram based ROUGE-2 and longest common subsequence based ROUGE-L F1 score respectively.
Our models also significantly outperform two related baseline methods HM-RNN and HO-RNN.

Furthermore, we compare the performance of MC-RNN with other recent methods on this task. We find that both ROUGE-1 and ROUGE-2 of our model outperform all current top systems, including \cite{shen2016neural,gehring2017convolutional,suzuki2017cutting}.

\begin{table}[t]
	\centering
	\small
	\caption{ROUGE F1 scores on Gigaword test set of abstractive summarization. RG-N stands for N-gram based ROUGE F1 score, RG-L stands for longest common subsequence based ROUGE F1 score. `$\dag$':  scores of Baseline-RNN. `$\star$': best scores of our method.}
	\begin{tabular}{l|r|rrr}
		\hline
		Methods   &Params& RG-1  & RG-2  & RG-L  \\ \hline
		\hline
		HM-RNN &35$M$&  34.68 & 16.11    & 32.22      \\ 
		HO-RNN &46$M$&   35.86 &  16.99& 33.38 \\
		Baseline-RNN &36$M$& \dag34.65  & \dag16.13 & \dag32.24 \\
 \hline
		
		MC-RNN-2&38$M$&36.21 & 17.30 & 33.60 \\
		MC-RNN-3&40$M$&$\star$36.55 & $\star$17.58 & $\star$33.72  \\
		MC-RNN-4&42$M$&36.50&  17.44& 33.68\\ \hline
		Gain from $\dag$ to $\star$ &-& +1.90 & +1.45 & +1.48       \\ \hline
	\end{tabular}
	\label{ABS:RNNsearch}
\end{table}

\subsection{Language Modeling}
\subsubsection{Experimental Setups}
We conduct our experiments on the Penn Treebank corpus which contains about 1 million words \cite{mikolov2010recurrent}, which has long been a central data set for experimenting with language modeling. We use perplexity as the evaluation metric.

MC-RNN uses LSTM as the recurrent unit on this task. Specifically, our network structures and regularization setups follow the state-of-the-art model, AWD-LSTM \cite{merity2017regularizing} on this task, using a stacked three-layer LSTM model, with 1150 units in the hidden layer and 400-dimensional word embeddings. DropConnect is used on the hidden-to-hidden weight matrices.

\subsubsection{Experimental Results}

\begin{table*}[!bt]
	\centering
	\caption{Single model perplexity on validation and test sets for the PTB language modeling task.}
	\begin{tabular}{l|rr}
		\hline
		Methods     & Validation     &Test   \\ \hline\hline
		Variational LSTM + augmented loss \cite{inan2016tying} & 71.1& 68.5 \\
		Variational RHN \cite{zilly2016recurrent}& 67.9 &65.4 \\
		NAS Cell \cite{zoph2016neural}& -& 62.4\\
		Skip Connection LSTM\cite{melis2017state}		& 60.9 & 58.3\\
		AWD-LSTM w/o finetune (baseline) \cite{merity2017regularizing}& 60.7& 58.8\\ \hline
		MC-RNN & 59.2 & 56.9\\  \hline
	\end{tabular}
	\label{LM}
\end{table*}

We compare perplexity of our model with other recent approaches in Table \ref{LM}. We take the recent top model, AWD-LSTM, as our baseline model.
To validate the effectiveness of our model itself rather than other factors, we report the performance of our model MC-RNN-3 without any post-processing tricks in the table.
Since the original AWD-LSTM baseline model is further improved by a model-specific optimization algorithm, as well as some other post-processing tricks, to focus on the performance of the model itself, we compare our method with the original ``AWD-LSTM w/o finetune''. This baseline directly reports the performance of the model without optimization tricks and post-processing tricks. From the table, we can find our method achieves improvement of 1.5 and 1.9 points perplexity on validation and test set respectively, compared to the baseline model.

Furthermore, by adopting the same post-processing tricks as AWD-LSTM (i.e., continuous cache pointer), MC-RNN even outperforms previously reported best result \cite{merity2017regularizing} of 53.9/52.8 on validation and test set, which uses optimization tricks while our model does not. MC-RNN achieves 53.3/52.6 perplexity, outperforming the state-of-the-art model by 0.6/0.2 on validation and test set.

\section{Analysis}
To get a deep insight of MC-RNN, in this section, we conduct case studies to investigate how the attention module helps the learning process. Moreover, we analyze the performance of our method with different sentence lengths.
\begin{figure*}[!t]
	\centering
	\begin{minipage}[t]{1\linewidth}
		\centering
		\includegraphics[page=3, width=1\linewidth]{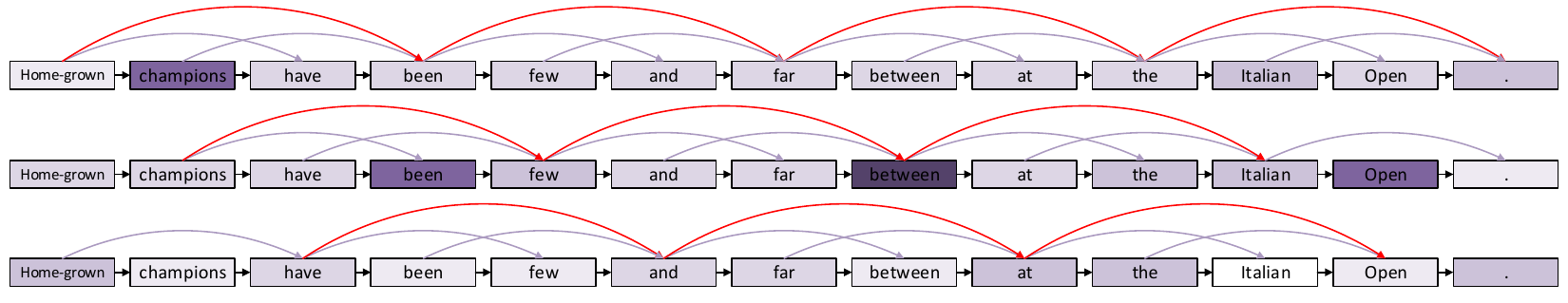}
	\end{minipage}~~~~~~~~~~~
    \vskip -0.5em
	\caption{Visualization of attention scores of the sentence `` \textit{Home-grown champions have been few and far between at the Italian Open.}" from Gigaword. Local dependence patterns and local structures are captured such as "home-grown champions", "champions have been", "few and far between", "Italian Open". }
    \vskip -0.5em
	\label{case}
\end{figure*}
\subsection{Case Studies and Visualization}
We conduct case studies to investigate what the model learns and how the model works. Figure \ref{case} presents the internal states of MC-RNN when processing a sentence from Gigaword test set of abstractive summarization task. We use the model with block size 4 and 3 channels. The darkness of the hidden nodes represents the value of attention scores. Specifically, the bigger the attention score, the darker the color.

From Figure \ref{case}, we can observe that the attention mechanism plays an important part in modeling local dependence. In particular, we list and analyze some dependence patterns of the sentence ``\textit{Home-grown champions have been few and far between at the Italian Open.}" (1) The $4$-$th$ attention score of channel 2 is significantly larger than those in other channels. Correspondingly, word ``been" is directly connected with the previous two words, together forming a subject-predicate structure, i.e., ``champions have been". (2) Another large attention score appears at the $8$-$th$ place at channel 2. The corresponding channel connects ``few and far between" together, which is a phrase meaning ``scarce and infrequent". (3) The last but one attention score of channel 2 represents a noun phrase, ``Italian Open'', which is an event name. (4) The 2-$nd$ score of channel 1 stands for ``home-grown champions''. These observations clearly demonstrate that MC-RNN can effectively model diverse local dependence patterns explicitly, including phrases and subject-predicate structures.


\subsection{Performance on Long Sentences}
\begin{figure}[t]
	\centering
	\includegraphics[page=8, width=0.9\linewidth]{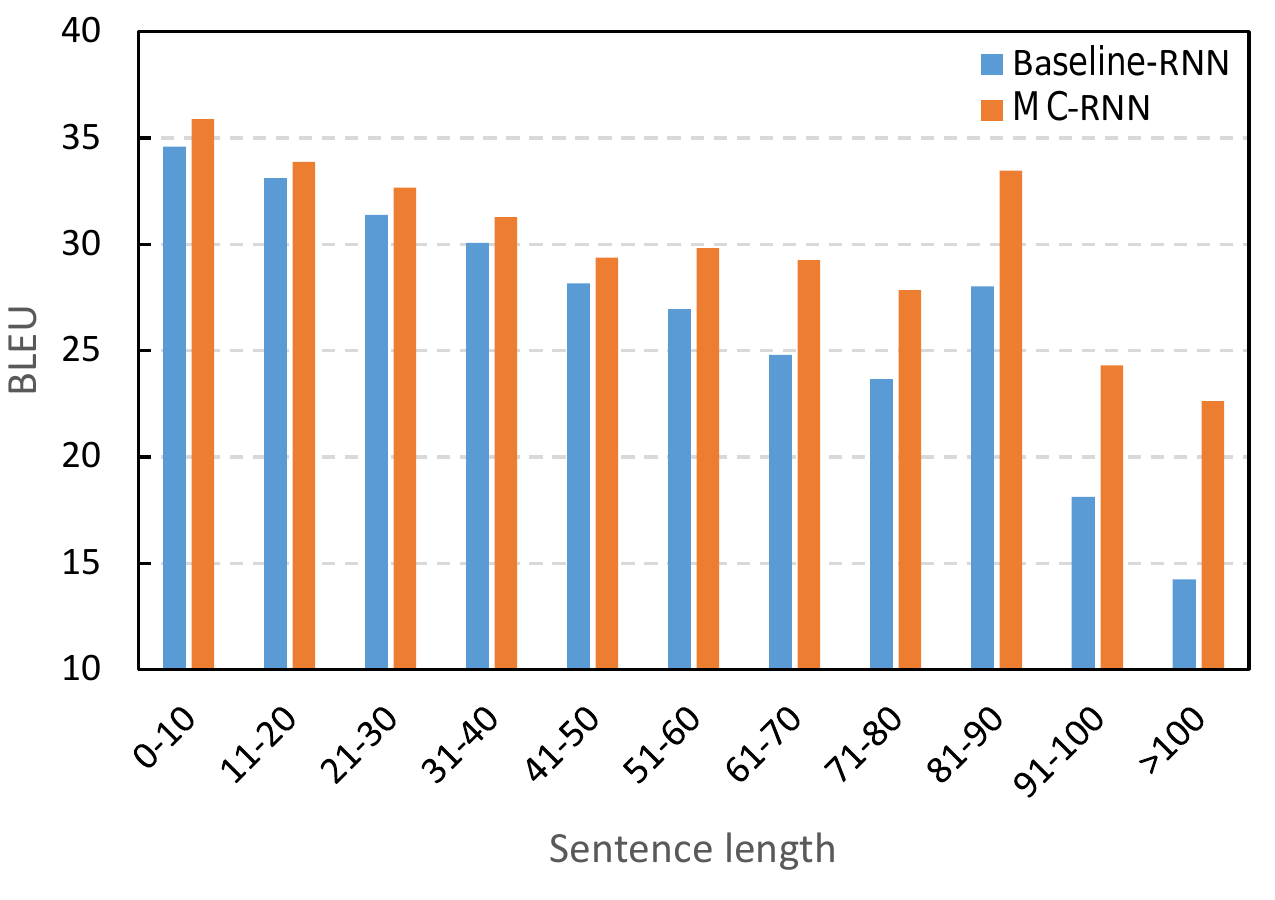}
	\vskip -1em
	\caption{BLEU scores of Baseline-RNN and MC-RNN on De-En translation task.}
	\label{fig:length}
	\vskip -0.5em
\end{figure}

We study the model performance on different lengths of sentences and observe an interesting phenomenon, that is our model works significantly better on long sentences than baselines.
Figure \ref{fig:length} shows the performance of our model compared to Baseline-RNN by different sentence lengths. Studies are conducted on IWSLT-14 De-En translation task, the same machine translation experiments as we described in previous section. We can observe that both our method and the baseline-RNN model perform worse as the lengths of the sentences increase, indicating long sentences are more difficult to handle than short ones. However, our model brings much more improvement on long sentences.
To be specific, when the sentence length is greater than 61, our model outperforms baselines by a larger margin (more than 4.0 BLEU scores) compared with the case when the sentence length is less than 40 (less than 1.0 BLEU score).

To understand why MC-RNN achieve better performance when handling long-term dependence in sequences, we make some theoretical analysis and try to give some insight. Following \citeauthor{chang2017dilated}, we define $d_i(l)$ as the length of the shortest path from node at time $i$ to node at $i + l$.  For conventional RNNs, $d_i(l) = l$, while for MC-RNN, $d \leq \lfloor \frac{l}{n-1} \rfloor  +1$, because when gradients are propagating backwards, the shortest paths to the nodes within a block are equal. MC-RNN enables short-cut connections across timestep and directly passes error signal through blocks. In this way, gradient vanishing can be alleviated.
Therefore, MC-RNN enables faster information flow and easier gradient propagation.

There is another work called DenseNet \cite{huang2016densely} which also use skip connections. Interestingly, we both enhance information flow by different means.
DenseNet applies skip connection among layers, ensuring gradient signal could be propagated without much loss from the end to the beginning of the network.
while MC-RNN  is different from DenseNet. MC-RNN applies skip connection among recurrent nodes in one layer. The skip connection in our MC-RNN can not only shorten the paths of gradient propagation but also explicitly model local dependence patterns of text sentences.

\subsection{Impact of Model Size and Time Cost}\label{timecost}

Since MC-RNN uses more parameters than Baseline-RNN, we also conduct experiments to exclude the improvement of performance caused by larger model size.
(1) We increase the size of the hidden state of Baseline-RNN model from 256 to 286. This model is called Baseline-RNN-large. (2) We increase the number of layers from 2 to 3 and we call it Baseline-RNN-deep. From Table \ref{MNT_MS} we can see that the model sizes of Baseline-RNN-large and Baseline-RNN-deep are almost the same as our best model, MC-RNN-3, while there is no significant improvement of performance.
These observations demonstrate that the better performance of our MC-RNN is caused by model design rather than larger model size.

In terms of computation, the proposed model is more expensive compared to conventional RNNs.
However, most parts of MC-RNN can be implemented in parallel, resulting in a practical time close to conventional RNN (i.e. O($n$) where $n$ is sentence length). First, different channels can be computed in parallel. Thus, the time cost of multiple channels is the same as that of one channel plus communication cost. Second, for time step $t$ within a single channel, i.e. Eqn. \ref{stk}, $s_{t-1}^k$ can also be implemented in $m_t^k$ parallel matrix multiplications of $W_{jm} h^k_{t-j}$, leading to similar time cost to one matrix multiplication. Thus, MC-RNN can achieve almost the same time cost as the conventional RNN. 

\begin{table}[t]
	\centering
	\caption{BLEU scores on IWLST 2014 De-En dataset with different model sizes.}
\vskip -0.5em
	\begin{tabular}{l|r|r}
		\hline
		Methods   & Params  & BLEU   \\ \hline\hline
		Baseline-RNN & 25$M$ & 31.03 \\
		Baseline-RNN-large & 29$M$ & 30.93 \\
		Baseline-RNN-deep & 29$M$ & 30.98\\ \hline
		MC-RNN-2 &  28$M$&31.98 \\
		MC-RNN-3 &  29$M$& $\star$32.23 \\
		MC-RNN-4 & 31$M$& 32.09 \\  \hline
	\end{tabular}
	\label{MNT_MS}
	\vskip -1em
\end{table}

\section{Conclusion and Future Work}

In this work, we proposed a new RNN model with multi-channel multi-block structure to better capture and utilize local patterns in sequential data for language-related tasks. Experiments on machine translation, abstractive summarization, and language modeling validated the effectiveness of the proposed model. We achieved new state-of-the-art results on Gigaword on text summarization and  Penn Treebank on language modeling.
For the future work, we will apply our model to more tasks, such as question answering, image captioning and so on.
\section{Acknowledgement}
We sincerely thank Tao Qin for his constructive suggestions during the work and his guidance on writing. 

This work is partially supported by NSF of China (61602266, 61872201, U1833114), Science and
Technology Development Plan of Tianjin (17JCYBJC15300, 16JCYBJC41900) and the Fundamental Research Funds for the Central Universities and SAFEA: Overseas Young Talents in Cultural and Educational Sector.

\bibliography{aaai2019}
\bibliographystyle{aaai}

\end{document}